# Emoji Prediction: Extensions and Benchmarking

Weicheng Ma
Dartmouth College
Weicheng.Ma.GR@dartmouth.edu

Ruibo Liu
Dartmouth College
Ruibo.Liu.GR@dartmouth.edu

Lili Wang
Dartmouth College
Lili.Wang.GR@dartmouth.edu

Soroush Vosoughi
Dartmouth College
soroush@dartmouth.edu

## ABSTRACT

Emojis are a succinct form of language which can express concrete meanings, emotions, and intentions. Emojis also carry signals that can be used to better understand communicative intent. They have become a ubiquitous part of our daily lives, making them an important part of understanding user-generated content. The emoji prediction task aims at predicting the proper set of emojis associated with a piece of text. Through emoji prediction, models can learn rich representations of the communicative intent of the written text. While existing research on the emoji prediction task focus on a small subset of emoji types closely related to certain emotions, this setting oversimplifies the task and wastes the expressive power of emojis. In this paper, we extend the existing setting of the emoji prediction task to include a richer set of emojis and to allow multi-label classification on the task. We propose novel models for multi-class and multi-label emoji prediction based on Transformer networks. We also construct multiple emoji prediction datasets from Twitter using heuristics. The BERT models achieve state-of-the-art performances on all our datasets under all the settings, with relative improvements of 27.21% to 236.36% in accuracy, 2.01% to 88.28% in top-5 accuracy and 65.19% to 346.79% in F-1 score, compared to the prior state-of-the-art. Our results demonstrate the efficacy of deep Transformer-based models on the emoji prediction task. We also release our datasets at https://github.com/hikari-NYU/Emoji_Prediction_Datasets_MMS for future researchers.

## 1 INTRODUCTION

Emojis are iconic tokens frequently used in natural language, especially in social media posts. Starting from symbolic expressions carrying emotional features (e.g. :) for smiling with joy), emojis have gradually grown to be a family of over 2,000 icons expressing emotions (e.g. 😃 for happiness), concrete semantic meanings (e.g. 🍱 for a meal), and intentions (e.g. 🎉 for celebrating). The combined use of emojis can express even more complex feelings, e.g. expressing sarcasm by attaching a smiley face to a discouraging message. Different from words, emojis are usually highly abstractive and are suitable for representing the stylistic features of a long span of text. Linking written text to emojis benefits the extraction of the abstract contextual information, e.g. sentiments, from the text. According to Na'aman et al. [6], emojis can serve as syntactic components in the text in the same way words do.

The emoji prediction task aims at finding the proper emojis associated with the text. It is an important natural language processing (NLP) task since the knowledge learned in the emoji prediction task can be well transferred to other tasks including emotion prediction, sentiment analysis, and sarcasm detection [4]. For example, detecting the sarcasm directly from the Twitter post "What a nice day. 😖" is difficult. But if we can correctly detect the emoji 😌strongly related to the content in the message, we will easily find the sarcasm lying in this post since the emojis 😌 and 😖 express right the opposite emotions. With the extended emoji-set, we observe much more potential of emoji prediction models in the NLP field. However, as the research on the emoji prediction task is still at an early stage, there are still many obstacles to its development.

The first problem is the availability and quality of the data. Emojis mainly appear in social media posts, e.g. tweets from Twitter. Most social media do not allow their data to be shared publicly (for various reasons, including privacy concerns). Corpora with social media contents are rare and often small in size (e.g. SemEval Twitter corpora [7, 9]). Most corpora do not release the actual posts from social media but rather links or ids pointing to them. However, these corpora become obsolete easily since social media users commonly modify or delete their posts. Almost all the existing research in this area is evaluated on individually collected datasets. This makes the model performances in the emoji prediction task incomparable and impedes the development in this field.

The quality of the annotations in the social media corpora is not guaranteed either. Manual annotation is not applicable on these datasets due to their large sizes (e.g. 1,246 million tweets used in training the DeepMoji model). Most researchers annotate the datasets with manually designed heuristics. Felbo et al. [4] implements this technique, extracting the emojis appearing in each tweet and using them as labels; tweets with multiple different emoji occurrences are duplicated with different labels. This technique, however, introduces noise to the datasets in the cases where there are input errors (e.g. a user wrongly clicks an emoji near the intended one) or when the emojis are used randomly, not connected to the content. Data imbalance is another key concern of the dataset quality when training deep learning models. The most frequently used emoji, 😂, appears five times more frequent than the second frequent emoji in the Gab posts related to the Charlottesville Event, for example [5]. The most common solution to this problem is by downsampling the data associated with the frequent emojis while upsampling those bound to rare emojis.

In addition to the lack of standard evaluation datasets, the label-set for the emoji prediction task is not fixed either. Felbo et al. [4] cluster the emojis appearing in their test dataset and use the 64 emoji types as labels. Barbieri et al. [2], instead, perform experiments on four label-sets containing 20, 50, 100 and 200 most frequent emojis. This is a more appropriate way of emoji-set construction to us, since the frequent emojis are better associated with the users' tweeting habits.

To address these problems, we clean up and label emoji prediction datasets consisting of Twitter posts to enable evaluations and comparisons across models in this paper. We create multiple datasets with different emoji-sets from the entire corpus. We also introduce a multi-label classification setting to the emoji prediction task to allow finer-grained evaluations. We hope that by re-defining the emoji prediction task and providing standard evaluation datasets, we will attract more research interest to the emoji prediction task.

Most existing research on the emoji prediction task uses RNNs (Recurrent Neural Networks) with the attention mechanism [2, 4, 8]. Barbieri et al. [1] incorporate visual information into the emoji prediction process. They base the emoji inference on images from Instagram [1] with the captions or descriptive texts. Nonetheless, these models often loses information when encoding long spans of text. It has become a trend to use the Transformer networks in the NLP community since last year. The Transformer-based models perform surprisingly well on a wide range of NLP tasks. Since we can easily gain a large volume of labeled data for the emoji prediction task using heuristics, evaluating the power of the Transformer networks on this task becomes a natural choice. We take advantage of the pre-trained BERT model [3] and fine-tune it on our datasets. We display the strength of BERT with both automatic and manual analysis under the two task settings. The evaluation results suggest that the BERT model outperforms the state-of-the-art emoji prediction model by large margins. This reveals the outstanding power of BERT in NLU tasks.

The BERT model performs well on the emoji prediction task, but the task is far from solved. By examining the error cases the BERT model generated, we still find noise in the annotations. As discussed above, this might have been caused by input errors or random usages. An alternative explanation is that each emoji might be summarized from a fraction of the text instead of the entire post.

[1] https://www.instagram.com

Our contributions in this paper are three-fold. First, we formally define the emoji prediction task under the multi-class and multi-label classification settings. Second, we annotate a large corpus from Twitter posts for the emoji prediction task and make it publicly available to benefit interested researchers. We construct evaluation datasets under two different settings. One of our datasets is built upon the 64 emoji-type setting applied by Felbo et al. [4]. The rest emoji sets are sampled from the frequent emojis in our corpus. Third, we construct a model based on pre-trained BERT and evaluate it on the datasets we create. The success of the BERT-based model reveals the power of the Transformer networks on the emoji prediction task and the other multimedia processing tasks as a whole. We display the evaluation results and manually analyze the predictions in later sections.

Over these years, emojis have been an important component in daily communications of human beings. We hope our research findings attract more research interest from the NLP community and thus push the research on multimedia processing forward.

## 2 THE TASK

The goal of the emoji prediction task is to predict the most appropriate emoji(s) given a piece of text. Most previous research regards the task as a multi-class classification problem. Inheriting from this, we add a multi-label classification setting to the task. We formally define the emoji task under the two formulations as follows. For clarity, we represent a document with $k$ words by $d = \{w_1, w_2, ..., w_k\}$. We refer to the emoji label-set with $E$ and use $e \in E$ to denote one emoji in the label-set.

**Multi-class Classification**: Given $d$, predict the $e \in E$ which best associates with $d$.

**Multi-label Classification**: Given $d$ and $E$, predict whether each $e \in E$ is properly connected to $d$.

Most existing research studies the emoji prediction task under the multi-class classification setting where the predictions are made by calculating and thresholding the probability distribution over $E$ given $d$. Different from the multi-class classification setting, the probability score of each emoji is independent from the rest emojis under the multi-label classification setting. Emojis with low probability scores are eliminated from the final predictions. This agrees better with the real scene where $d$ is not always associated with a fixed number of emojis. Meanwhile, it is difficult to approach the multi-label emoji classification task with the multi-class setting since the number of emojis is large and as the amount of emojis to choose for each tweet is not fixed. Despite the increased difficulty than the multi-class classification setting, the predictions of the BERT model under the multi-label classification formulation agrees well with the labels in our experiments.

## 3 DATASETS

There has not been a publicly available emoji prediction dataset yet, and the choice of $E$ has never reached an agreement. This impedes the development of research on the emoji prediction task since the lack of a benchmark dataset and a shared emoji label-set makes it difficult and unfair to compare the performances of emoji prediction models. In this paper, we construct emoji prediction datasets out of the Celebrity Profiling corpus released by Wiegmann et al. [10]

| Tweet | Label |
|---|---|
| Incredibly moving and authentic Congrats friends @ladygaga #bradleycooper @ New York, New York https://t.co/WZsrZuPAeh | ❤️ |
| Happy #PRIDE month. I see you and I love you | ❤️💛💚💙💜🌈 |

**Table 1: Two example tweets and their respective labels in our dataset.**

| Dataset | Size | Avg. #Tokens | Vocab Size |
|---|---|---|---|
| ML | 1,480,685 | 19.44 | 2,445,157 |
| MC-20 | 180,660 | 17.92 | 434,214 |
| MC-50 | 455,422 | 18.43 | 911,545 |
| MC-64 | 564,167 | 19.37 | 1,125,338 |
| MC-100 | 921,341 | 19.27 | 1,649,966 |
| MC-150 | 1,389,870 | 19.28 | 2,244,426 |
| MC-200 | 1,858,741 | 19.39 | 2,749,149 |
| MC-250 | 2,254,348 | 20.02 | 3,203,071 |
| MC-300 | 2,548,399 | 20.46 | 3,523,353 |

**Table 2: An analysis of the datasets construct by us. ML refers to the multi-label classification dataset while the rest datasets are for the multi-class classification setting. The size column denotes the number of records in each dataset. Avg. #tokens refers to the average number of tokens per sentences in the datasets and the vocab size is the number of unique tokens in the datasets.**

to enable evaluations on this task. The Celebrity Profiling corpus contains tweets posted by 48,335 verified accounts on Twitter, so the contents are generally high in quality and relatively formal in language usage. Additionally, since the birth years of these authors range from 1940 to 2012, the corpus shows no bias to the age-specific habits of emoji usage. Two sample tweets from the corpus are shown in Table 1, the first of which is under the multi-class setting and the second is under multi-label setting. As for the selection of the label-set, the majority of research on the emoji prediction task to date chooses to use the most frequent emojis in the dataset [1] or a handcrafted emoji-set [4]. It is the merit of the manually engineered emoji-set that the emojis usually carry strong emotional or concrete meanings. This benefits the prediction process but loses generality in the analysis of human blogging patterns. In comparison, choosing the emoji-set by frequency compensates for social behavior analysis but adds difficulty to the prediction task. We combine the two strategies by first structuring the emoji-set with frequent emojis in the dataset and then sanitizing the emoji list manually. From the full set of 2,811 emojis, we select the most frequent 300 emojis to construct our label-sets. The least frequent emoji in our emoji list is bound to 5,704 tweets in the corpus.

We annotate the dataset with heuristics. The emojis, after the annotation process, are eliminated from the content and empty tweets are cleared from the resulted dataset. We get rid of the emojis not appearing in our label-set and randomly sample 20% of the data to form our multi-label classification dataset. The sampling process follows the original distribution of the label counts in the dataset. In the pre-processing step, we remove sentences less than three words long to get rid of noisy contents. Though this removes the meaningful contents including "Happy Birthday", it makes more sense to apply the BERT-based model on complicated scenes. The overly simple sentences can be handled by much simpler models. The final multi-label classification dataset contains 1,480,685 records, with an average number of 1.89 emojis per tweet. For the multi-class classification setting, we first duplicate the tweets with multiple labels and assign each of them one single emoji, as Felbo et al. [4] do. To avoid bias under the multi-class classification setting, we downsample the tweets associated with the overly frequent emojis in the dataset. We set the maximum number of tweets per emoji to 10,000 and construct a multi-class classification dataset containing 2,548,399 records after normalization. The sizes and average sentence lengths of our datasets are reflected in Table 2. To validate the appropriateness of using different emoji-sets in the prediction task, we group the tweets by their labels in the multi-class classification dataset and divide the dataset to multiple subsets by choosing the tweets with the most frequent 20, 50, 100, 150, 200, 250, and 300 emojis. For comparability, we also create a dataset with the 64 emoji types used by Felbo et al. [4]. All the datasets are partitioned into train/dev/test sets with 80%/10%/10% of the entire data respectively, with 29936 as the random seed.

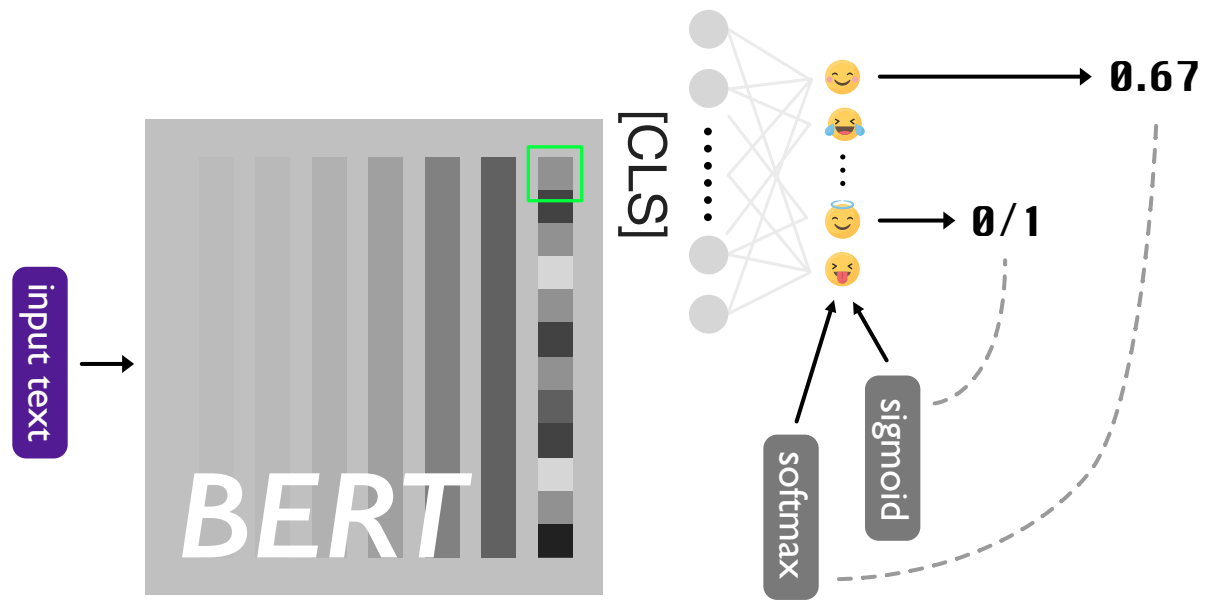

**Figure 1: The BERT model architecture. The multi-class classification model applies the softmax activation function on top of the dense layer while the multi-label classification uses the sigmoid activation function.**

In the later sections, we refer to the multi-label classification dataset by *ML* and the multi-class classification datasets using their respective size of the label-set, e.g.*MC-20* for the dataset with the top 20 frequent emojis.

## 4 MODEL ARCHITECTURE

Preceding research has proven the efficacy of bidirectional contextual information and the attention mechanism on the emoji prediction task. Regarding the recent success of pre-trained Transformer-based models on multiple NLP tasks, we apply the BERT model on the emoji prediction task. The core of the Transformer networks is the multi-layer self-attention and the positional encoding. Merited from its bidirectional nature and the large pre-training corpus, BERT shows outstanding potential in understanding natural language. We fine-tune a pre-trained BERT model on our dataset to adapt BERT to the social media domain. On top of the BERT model, we stack a single linear layer and a softmax layer to scale the dimension of the BERT output down to the prediction space. We train the model with cross-entropy loss, comparing the predicted probability distributions with the one-hot labels. Under the multi-class classification setting, the prediction space is of size $|E|$ where $|\cdot|$ calculates the size of a set. We use softmax activation under this setting and rank the emojis according to their probability scores at the prediction stage. For the multi-label classification setting, we shape the prediction space as $|E| \times 2$ and apply sigmoid application on it. Each $e$ is present in the prediction if its probability score after the sigmoid activation is greater than 0.5. The architecture of our multi-class and multi-label classification models is shown in Figure 1.

| ID | Content | Label | Prediction |
|---|---|---|---|
| 1 | good night bonne nuit la toile tomorrow is a new day ! Celebrate, believe, achieve! https://t.co/7gxor4FMfx | 😘 | 🌙 |
| 2 | luk happens wen preparation meets opportunity...love happens wen i meet u. | © | ❤️ |
| 3 | we animals is doing big things! become a supporter today help us take on more projects to help animals t.co/pfnumadnumznumb t.co/lsetyxxymk | ❤️ | 💚 |
| 4 | d12 new mix tape.. coming soon!! we did it 24 hours..!!!https://t.co/sPEdgfM6LE | 👕 | 🔥 |
| 5 | rt @peterjv26: @free_thinker @vivekagnihotri yes sure. i am in #MeTooUrbanNaxal | 🙏 | 🙏 |
| 6 | @rankinphoto Me too one of my faves | ❤️ | ❤️ |

**Table 3: Example predictions made by our BERT model.**

| Dataset | Model | ACC | ACC@5 | F-1 |
|---|---|---|---|---|
| MC-20 | DeepMoji | 42.11 | 74.68 | 30.51 |
| | BERT | **54.65** | **76.18** | **54.70** |
| MC-50 | DeepMoji | 23.50 | 51.69 | 19.91 |
| | BERT | **43.08** | **63.53** | **42.89** |
| MC-64 | DeepMoji | 23.36 | 49.41 | 19.03 |
| | BERT | **41.88** | **61.95** | **41.44** |
| MC-100 | DeepMoji | 23.16 | 46.93 | 17.42 |
| | BERT | **77.90** | **88.36** | **77.83** |
| MC-150 | DeepMoji | 21.97 | 45.13 | 16.58 |
| | BERT | **38.33** | **57.84** | **37.84** |
| MC-200 | DeepMoji | 21.13 | 42.51 | 16.06 |
| | BERT | **38.48** | **57.68** | **38.02** |
| MC-250 | DeepMoji | 20.66 | 40.17 | 14.31 |
| | BERT | **38.31** | **57.42** | **38.07** |
| MC-300 | DeepMoji | 20.19 | 34.89 | 12.97 |
| | BERT | **46.66** | **62.07** | **46.73** |

**Table 4: The experimental results on eight multi-class classification datasets. BERT refers to the BERT model. The best performances on each dataset are in bold. The three metrics we use in the evaluations are ACC (Accuracy), ACC@5 (Top 5 Accuracy) and F-1 score.**

| Content | DeepMoji | BERT | Label |
|---|---|---|---|
| even when i m early i m late | 😅😓😑😪😐 | 😞😫💔😷💀 | 😷 |
| katkingsley we could learn that in a day for a friday afternoon workshop though mrs. t!?! | 😜😉👍😄😳 | 😁😀🎵😍😋 | 😁 |

**Table 5: The top 5 outputs of the DeepMoji model and the BERT model on two example sentences.**

## 5 EXPERIMENTS AND ANALYSIS

We use the pre-trained **bert-base-cased** model release by Google in our evaluations. The model contains 12 Transformer encoder layers, each of which consists of 12 attention heads. The hidden size of the model is 768. In the experiments, we fine-tune the BERT model on our dataset for 5 epochs under both the multi-class and multi-label classification settings, with 0.0001 as the learning rate. The longest message contains 128 words after tokenization, so we set the maximum sequence length to 128 in our experiments. We apply a batch size of 64 in the evaluations. The dataset is randomly shuffled before use. We use the DeepMoji model as our baseline. The model is pre-trained on a Twitter corpus containing 1,246 million tweets and is fine-tuned on each of our training datasets for 4 epochs before evaluating. We use F-1 score to evaluate the models under the multi-class classification setting. Under the multi-label classification setting, we instead group the data points by the emojis and calculate accuracy score for each emoji.

We display the experimental results under the multi-class classification setting in Table 4. On all the eight datasets, the BERT model outperforms the DeepMoji model by large margins, though the DeepMoji model is pre-trained on a much larger dataset. This demonstrates the superior encoding ability of the pre-trained BERT model. The DeepMoji model suffers from a noticeable performance drop with the growth of the label space, while the BERT model generalizes well to the more complex problem settings. We display 2 sample outputs of both the BERT model and the DeepMoji model in Table 5. In both examples, the BERT model correctly captures the overall emotions in these tweets. In the first sentence, though the labeled emoji is not the top-ranked one in our predictions, all the predicted emojis are associated with negative or disappointed emotions. The BERT model well addresses the positive sentiment in the second sentence as well, generating the smiling, lovable faces as the output. The DeepMoji model also performs well on the second sentence, except for the "flushed face" emoji at the fifth place in its output. On the first tweet, however, the DeepMoji model outputs two "sweat face" emojis, two "emotionless face" emojis and one

| | | | | | | | | | | | | | |
|---|---|---|---|---|---|---|---|---|---|---|---|---|---|
| ❤️ | 😂 | 👍 | 🙏 | 🙌 | 😘 | 😍 | 😊 | 🔥 | 👏 | 👌 | 💪 | 👊 | 😉 |
| 89.18 | 93.04 | 94.55 | 95.72 | 95.11 | 95.93 | 95.72 | 96.32 | 96.97 | 96.48 | 96.84 | 96.88 | 97.20 | 97.56 |
| 🎉 | 😎 | 😁 | 💯 | 😜 | 👀 | 💕 | ☺️ | ✨ | ✌️ | ⚽ | 😳 | 🙈 | 😭 |
| 97.28 | 97.71 | 97.83 | 98.21 | 98.14 | 98.53 | 98.08 | 98.59 | 98.12 | 98.41 | 98.56 | 98.65 | 98.45 | 98.57 |
| 💙 | 💋 | 🤔 | 🎶 | 😀 | 🤣 | 💃 | 💜 | 🤗 | 😄 | 😃 | ✊ | 😩 | ☀️ |
| 98.40 | 98.54 | 98.98 | 98.38 | 98.87 | 98.98 | 98.51 | 98.75 | 98.96 | 98.98 | 98.95 | 99.17 | 98.97 | 98.91 |
| 💥 | 🏆 | ❤️ | 💛 | 😏 | 💖 | 🤘 | 😱 | 💚 | 👇 | 😬 | ♥️ | 🤷 | 🎂 |
| 98.84 | 98.99 | 99.21 | 98.96 | 99.21 | 99.03 | 99.20 | 99.14 | 99.03 | 99.42 | 99.23 | 99.21 | 99.33 | 98.93 |
| 👉 | 📷 | 🙄 | 😆 | ⚡ | 🏀 | 😝 | 📸 | 🏈 | 🎥 | 🎈 | 🌹 | 🎤 | 👑 |
| 99.28 | 99.35 | 99.42 | 99.35 | 99.32 | 99.39 | 99.32 | 99.34 | 99.38 | 99.31 | 99.07 | 99.37 | 99.16 | 99.36 |
| 🎄 | 🤦 | 😢 | 😅 | 🎁 | 🎵 | 😈 | 😋 | 🌟 | 😔 | ⭐ | ‼️ | 🐶 | ❄️ |
| 99.49 | 99.53 | 99.52 | 99.44 | 99.12 | 99.94 | 99.48 | 99.49 | 99.36 | 99.61 | 99.46 | 99.52 | 99.54 | 99.60 |
| 🌈 | 👋 | 🖤 | 😇 | 🙊 | 😴 | 💔 | 🙋 | 💁 | 😒 | 📺 | 😡 | 🏃 | 💀 |
| 99.51 | 99.66 | 99.60 | 99.55 | 99.53 | 99.64 | 99.66 | 99.6 | 99.56 | 99.67 | 99.64 | 99.68 | 99.55 | 99.65 |
| 🍾 | ☕ | 💫 | 🎊 | ✅ | 🌴 | ✔️ | 💨 | 🍻 | 🚀 | 🎧 | 🤓 | 👯 | 👻 |
| 99.43 | 99.66 | 99.54 | 99.35 | 99.66 | 99.57 | 99.67 | 99.59 | 99.60 | 99.68 | 99.58 | 99.67 | 99.58 | 99.69 |
| 🎼 | 🍀 | 👍🏻 | 💞 | 🌊 | 🙂 | 🙃 | 🎅 | 🎵 | ⚾ | 🎬 | 🌞 | 😌 | 💰 |
| 99.51 | 99.67 | 99.73 | 99.65 | 99.65 | 99.76 | 99.73 | 99.64 | 99.55 | 99.72 | 99.62 | 99.66 | 99.72 | 99.68 |
| 😫 | 🌎 | 💦 | 🌸 | ⚪ | 🔵 | 😤 | ☝️ | 🕺 | 😑 | 🏉 | 🍷 | 🎸 | ⬇️ |
| 99.71 | 99.67 | 99.67 | 99.69 | 99.71 | 99.70 | 99.75 | 99.74 | 99.68 | 99.8 | 99.71 | 99.69 | 99.64 | 99.78 |
| 👎 | 🔊 | 😻 | 🏁 | 😛 | 😐 | 😞 | 🐐 | 🎃 | 💩 | 💘 | ⛳ | 🗣️ | 🥊 |
| 99.81 | 99.73 | 99.76 | 99.78 | 99.78 | 99.83 | 99.82 | 99.86 | 99.83 | 99.81 | 99.75 | 99.80 | 99.82 | 99.77 |
| 📻 | 👶 | 💓 | 💐 | 📚 | 🎾 | 😕 | ☔ | 💎 | 🏊 | 🏴 | 🌺 | 👨 | 🥂 |
| 99.81 | 99.74 | 99.76 | 99.74 | 99.77 | 99.82 | 99.85 | 99.82 | 99.78 | 99.76 | 99.80 | 99.76 | 99.74 | 99.74 |
| 😮 | 👩 | 🍰 | 🔫 | 😚 | 👸 | 🥇 | 💤 | 🦁 | 😷 | 🍺 | 🤞 | 💅 | ▶️ |
| 99.85 | 99.72 | 99.75 | 99.81 | 99.85 | 99.76 | 99.80 | 99.83 | 99.79 | 99.83 | 99.77 | 99.85 | 99.80 | 99.83 |
| ❗ | 🐾 | ✋ | ⚫ | 🤪 | 💝 | 🐝 | 🌍 | 🎯 | 💍 | 💣 | ™️ | 🐻 | 🍸 |
| 99.85 | 99.80 | 99.85 | 99.80 | 99.83 | 99.80 | 99.85 | 99.82 | 99.86 | 99.81 | 99.81 | 99.89 | 99.83 | 99.76 |
| 🎙️ | 😵 | 🍕 | 🎹 | 🍁 | 😥 | 🤩 | 💄 | 🦄 | 🚗 | 🚴 | 🌻 | 📽️ | 👽 |
| 99.80 | 99.86 | 99.85 | 99.76 | 99.86 | 99.89 | 99.84 | 99.81 | 99.82 | 99.84 | 99.82 | 99.80 | 99.84 | 99.87 |
| 🏅 | 🙅 | 🍴 | ♡ | 🏏 | 🔝 | ⛄ | 🌷 | 🌲 | 🙆 | 👠 | ⏰ | ☘️ | 📝 |
| 99.85 | 99.86 | 99.84 | 99.90 | 99.87 | 99.83 | 99.83 | 99.85 | 99.85 | 99.85 | 99.81 | 99.85 | 99.86 | 99.87 |
| 🎟️ | 📍 | 🏌️ | 🐰 | 😪 | 👆 | 😓 | 🧡 | 🦋 | 😰 | 🍊 | 😵 | ❣️ | 😖 |
| 99.86 | 99.88 | 99.87 | 99.89 | 99.88 | 99.88 | 99.91 | 99.85 | 99.87 | 99.87 | 99.88 | 99.89 | 99.87 | 99.89 |
| 🐯 | 💻 | 🎭 | 🙉 | 🌙 | ☠️ | ❌ | 🎩 | 📰 | 💭 | 🎨 | 📱 | 🍹 | 👼 |
| 99.87 | 99.84 | 99.86 | 99.87 | 99.88 | 99.88 | 99.89 | 99.88 | 99.91 | 99.90 | 99.87 | 99.86 | 99.83 | 99.88 |
| 👈 | 👭 | 🍦 | 🎀 | 🏳️ | 🦊 | 🔪 | ®️ | 👙 | 🍔 | 👅 | 🏄 | 👧 | 😙 |
| 99.88 | 99.86 | 99.87 | 99.87 | 99.87 | 99.89 | 99.88 | 99.91 | 99.83 | 99.88 | 99.89 | 99.89 | 99.85 | 99.90 |
| 🐱 | 💸 | 🍑 | 🔑 | 🐸 | 😶 | 🧐 | 🍂 | 🌝 | 🍎 | 😣 | 🌿 | 📹 | 🐊 |
| 99.89 | 99.90 | 99.90 | 99.93 | 99.89 | 99.93 | 99.93 | 99.90 | 99.93 | 99.90 | 99.91 | 99.89 | 99.91 | 99.91 |
| 🌅 | 🗽 | 👗 | 💡 | 🌏 | 🏒 | 🤤 | 🍳 | 🐍 | 👐 | 🤐 | 🙀 | 🤫 | 👫 |
| 99.91 | 99.90 | 99.85 | 99.91 | 99.89 | 99.9 | 99.92 | 99.88 | 99.91 | 99.91 | 99.94 | 99.92 | 99.93 | 99.89 |
| 🙇 | 🏇 | 📖 | 😨 | 👁️ | 🐴 | | | | | | | | |
| 99.90 | 99.91 | 99.89 | 99.92 | 99.93 | 99.92 | | | | | | | | |

**Table 6: The accuracy score for each emoji (sorted by frequency) under the multi-label classification setting. The average accuracy score across all the 300 emojis is 99.41%.**

"sleepy" emoji, which are not related to the content. In general, the DeepMoji model predicts emojis with high accuracy on sentences with explicit emotional expressions but performs poorer than the BERT model on the sentences where the emotions are implied. This suggests that the BERT model is stronger on understanding the stylistic features than recurrent neural networks, agreeing with our evaluation results.

In Table 3 we show some additional predictions made by the BERT model. From the results, we observe the problem caused by the use of other languages than English in a portion of tweets. Sentence 1 in Table 3 contains French, for example. Since the BERT model we use is trained on an English corpus, expressions from other languages often act as noise and lead to prediction errors. Sentence 2 displays another type of tweet that tends to fool the BERT model. Though the tweet is written in English, its use of informal abbreviations (e.g., wen for when) makes it difficult for the BERT model to understand the content. About 19.62% of the tweets in our datasets either uses English informally or are written in foreign languages, according to our study. These records largely contribute to the total error cases in our experiments. Another type of error is caused by the diversity of emojis with similar meanings. Sentence 3, for example, is labeled with the "red heart" emoji and the BERT model predicts the "green heart" emoji instead. In the datasets, these emojis are often used interchangeably, making it difficult for the BERT model to learn the differences inside these emoji families. This problem frequently happens because people seldom notice the minor differences from the small icons. This can be addressed in extensions to this work by grouping the emojis with similar meanings into groups to weaken the inner-group inconsistency. It is worth noting that some labels in our datasets look unreasonable to us. Sentence 4 talks about a new tape while the label is a "T-shirt" emoji. We guess that some twitter users use the emojis randomly, without considering the actual meanings of their posts. These annotations harm the performance of the BERT model. Unfortunately, we are unable to manually check the qualities of all the annotations in our datasets. At this stage, we leave these improper annotations as noise in the datasets. An interesting characteristic of the social media data is the mixture of multiple modalities in a single post. Barbieri et al. [1] prove by experiments that visual information helps in the emoji prediction task. Based on our observation, the links and hashtags in the tweets also provide strong clues to the predictions. Resolving the hashtags to only their text parts weakens the clues. The hashtag "#MeToo" in Sentence 5 leads to the "folded hands" emoji, for instance, while the phrase "me too" in Sentence 6 does not. An extension of this work could take into account these additional information for prediction.

| | ❤️ | 😂 | 👍 | 🙏 | 🙌 |
|---|---|---|---|---|---|
| MC-20 | 50.05 | 55.02 | 53.26 | 54.58 | 52.75 |
| MC-50 | 28.07 | 38.84 | 43.04 | 49.74 | 37.34 |
| MC-64 | 52.42 | 92.59 | 45.89 | 46.80 | 47.77 |
| MC-100 | 61.09 | 75.56 | 78.67 | 73.84 | 73.62 |
| MC-150 | 10.51 | 27.65 | 31.34 | 35.89 | 24.21 |
| MC-200 | 14.46 | 27.41 | 30.37 | 32.79 | 22.36 |
| MC-250 | 12.95 | 25.80 | 29.85 | 31.35 | 21.59 |
| MC-300 | 36.11 | 44.08 | 41.22 | 43.92 | 41.73 |
| | 😘 | 😍 | 😊 | 🔥 | 👏 |
| MC-20 | 50.18 | 56.74 | 56.32 | 59.78 | 54.28 |
| MC-50 | 40.45 | 41.54 | 36.02 | 53.58 | 46.16 |
| MC-64 | 32.18 | 53.19 | 45.93 | 45.98 | 55.78 |
| MC-100 | 74.86 | 80.22 | 68.11 | 82.47 | 79.96 |
| MC-150 | 24.30 | 30.44 | 21.12 | 42.54 | 35.41 |
| MC-200 | 19.11 | 30.47 | 23.03 | 37.23 | 36.15 |
| MC-250 | 20.04 | 27.38 | 22.46 | 36.87 | 33.05 |
| MC-300 | 34.25 | 41.48 | 37.24 | 44.94 | 45.37 |

**Table 7: The performances of the BERT model on the top 10 most frequent emojis in all the experiments. We use the dataset names to denote the evaluations performed on these datasets.**

We also compare the performances of the BERT model on the top 10 most frequent emojis across the eight datasets for multi-class classification. The results are displayed in Table 7. Under different settings, the BERT model performs very differently on the emojis which appear in all the datasets, yielding the great influence of noisy data in terms of the frequent emojis. As the contents are consistent in style, it is highly possible that the extended emoji-sets contain similar emojis to the top-ranked ones. For example, the emojis "green heart", "purple heart" and "yellow heart" appearing after the 50th place might have caused the low performance of the BERT model on predicting the "red heart" emoji. We observe that the BERT model performs relatively stable on the 🔥 emoji. Our hypothesis is that the BERT model tend to be less influenced by the tweets bound to the emojis with less counterparts across datasets. Further experiments are needed to validate these assumptions.

Interestingly, the top 10 most frequent emojis all bear emotional meanings. This lends solid support to our assumption that people usually use emojis to express their feelings or emotions. The high performance of the BERT model on these emojis show that the BERT model is prominent in capturing emotional expressions. Predicting the emojis with concrete meanings (e.g., 🏡 for a house) is more difficult due to two reasons. First, different from the emotional emojis which abstracts the content, in most cases this type of emojis are used as semantic placeholders in a tweet. The emojis are eliminated from the content in our experiments, making it difficult to infer the emoji out of its context. Secondly, the emojis referring to objects are sometimes used randomly in Twitter posts. This introduces noise to the training datasets and thus harms the performance of the BERT model. A clean Twitter corpus is needed for more accurate predictions by removing the noisy emojis from the tweets.

The merit of the multi-label classification setting is that the irrelevant emojis are kicked out of the predictions. In our multi-label classification dataset, for example, the average number of emojis per tweet is 1.89. It is thus often problematic to represent the text with the top 5 most possible emojis, as Felbo et al. [4] do. We evaluate the BERT model under the multi-label classification setting and show the results in Table 6. The BERT model performs surprisingly well for all the emojis, revealing the fact that the dataset is overly simple for our multi-label emoji prediction model. Since the annotations come from the actual appearance of emojis in the tweets, the quality and completeness of the annotations are not guaranteed. In many cases, only one or two most probable emojis are selected as the label, while some tweets contain over 60 different emojis. The discreteness of the emojis is also a problem since similar emojis are regularly used together or interchangeably. This can be addressed by grouping similar emojis together.

## 6 CONCLUSION AND FUTURE WORK

The research on the emoji prediction task is relatively young in the NLP community. The task definition is vague, and no standard evaluation dataset exists for researchers to use. In this paper, we reformulated the task by formally defining two settings of the emoji prediction task. We also annotated several datasets based on Twitter posts with multiple sets of emojis as labels, for evaluating emoji prediction models. The emojis-sets were either handcrafted or selected from the most frequent emojis in the Twitter corpus with different thresholds. We benchmarked the datasets with both the DeepMoji model and the BERT model based on a pre-trained BERT. From the evaluation results, we found that the BERT-based model largely outperformed the DeepMoji model under both the multi-class and multi-label classification settings. This demonstrates the extensive power of pre-trained Transformer-based models on the emoji prediction task and potentially on other multimedia processing tasks. As a next step, we propose to expand the emoji prediction task to a more fine-grained, aspect-based classification setting, since the different emojis bound to one tweet tend to be correspondent to different parts of the content. On the other hand, by analyzing the predictions and errors made by the BERT model, we noticed that there are still flaws in the annotations in our datasets. Possible input misses and randomness in emoji choices were the two most common problems our datasets faced. Future work could also focus on further refinement of the annotations both on the tweet level and on the aspect level.

To aid reproducibility and future research, the data and code for this paper will be made available upon request.